\documentclass[10pt,twocolumn,letterpaper]{article}

\usepackage[final,applications]{wacv}      
\usepackage{amsfonts}
\usepackage{booktabs} 
\usepackage{multirow}
\usepackage{colortbl}

\definecolor{wacvblue}{rgb}{0.21,0.49,0.74}
\usepackage[pagebackref,breaklinks,colorlinks,allcolors=wacvblue]{hyperref}

\def\wacvPaperID{2599} 
\def\confName{WACV}
\def\confYear{2026}

\title{EVCC: Enhanced Vision Transformer-ConvNeXt-CoAtNet Fusion for Classification}

\author{
Kazi Reyazul Hasan, Md Nafiu Rahman, Wasif Jalal, Sadif Ahmed, \\
Shahriar Raj, Mubasshira Musarrat, Muhammad Abdullah Adnan\thanks{Authors are listed in order of contribution.\\Corresponding author. Email: adnan@cse.buet.ac.bd}\\
\\
Department of Computer Science and Engineering\\
Bangladesh University of Engineering and Technology (BUET)\\
Dhaka, Bangladesh
}

\begin{document}
\maketitle
\begin{abstract}
Hybrid vision architectures combining Transformers and CNNs have significantly advanced image classification, but they usually do so at significant computational cost. We introduce EVCC (\textbf{E}nhanced \textbf{V}ision Transformer-\textbf{C}onvNeXt-\textbf{C}oAtNet), a novel multi-branch architecture integrating the Vision Transformer, lightweight ConvNeXt, and CoAtNet through key innovations: (1) adaptive token pruning with information preservation, (2) gated bidirectional cross-attention for enhanced feature refinement, (3) auxiliary classification heads for multi-task learning, and (4) a dynamic router gate employing context-aware confidence-driven weighting. Experiments across the CIFAR-100, Tobacco3482, CelebA, and Brain Cancer datasets demonstrate EVCC's superiority over powerful models like DeiT-Base, MaxViT-Base, and CrossViT-Base by consistently achieving state-of-the-art accuracy with improvements of up to 2 percentage points, while reducing FLOPs by 25–35\%. Our adaptive architecture adjusts computational demands to deployment needs by dynamically reducing token count, efficiently balancing the accuracy-efficiency trade-off while combining global context, local details, and hierarchical features for real-world applications. The source code of our implementation is available at \url{https://anonymous.4open.science/r/EVCC}.
\end{abstract}

\section{Introduction}
\label{sec:introduction}

Image classification has long been a cornerstone of computer vision, serving as a gateway to more complex tasks such as object detection, semantic segmentation, and visual reasoning. Traditional convolutional neural networks (CNNs) such as ResNet~\cite{he2015deepresiduallearningimage} and EfficientNet~\cite{tan2020efficientnetrethinkingmodelscaling} have demonstrated exceptional performance in extracting hierarchical spatial features, utilizing locality and spatial invariance. However, recent advances in Transformer architectures have introduced a new paradigm that leverages self-attention for capturing long-range dependencies and global context. While Transformers have shown remarkable success in tasks like natural language processing and vision, they often lack spatial locality inductive biases, which can lead to data inefficiency, especially in low-data regimes.

Despite these limitations, the combination of CNNs and Transformers presents an opportunity to harness the complementary strengths of both architectures. CNNs are adept at capturing local patterns and spatial hierarchies, whereas Transformers excel in global context modeling through self-attention. However, naïvely integrating these architectures may result in redundant computations and increased parameter overhead.

Moreover, the demand for computationally efficient models has surged due to the proliferation of edge devices and resource-constrained environments. This necessitates the development of architectures that not only maintain high accuracy but also effectively balance computational cost and inference speed. The challenge lies in strategically fusing the representational power of CNNs and Transformers while mitigating their respective drawbacks, such as the overemphasis on locality in CNNs and the potential over-reliance on global context in Transformers.

\begin{figure*}[!htbp]
   \centering
   \includegraphics[width=1\linewidth]{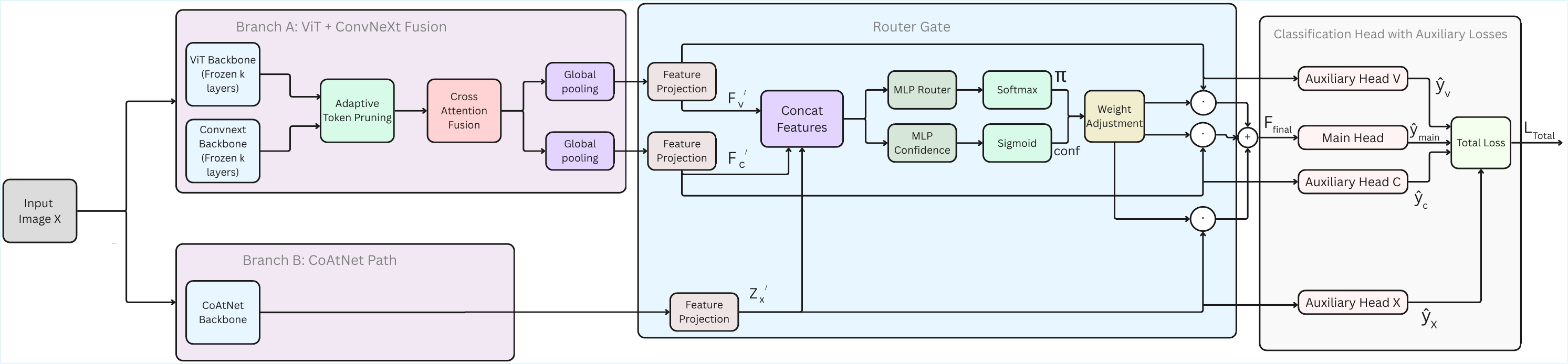}
   \caption{Overall pipeline of the EVCC architecture}
   \label{fig:pipeline}
\end{figure*}

To address these challenges, we propose \textit{EVCC} (see Figure \ref{fig:pipeline}, a novel multi-branch vision architecture that strategically combines a Vision Transformer (ViT) \cite{dosovitskiy2021image}, an improved lightweight ConvNeXt \cite{liu2022convnet2020s}, and a third auxiliary CoAtNet backbone. The core design philosophy of EVCC is to facilitate effective multi-stage fusion through a series of well-structured stages, enabling synergistic learning across the branches while maintaining computational efficiency.

The proposed architecture leverages three key modules to achieve this objective:

\begin{itemize}
  \item \textbf{Adaptive Token Pruning:} Both the ViT and ConvNeXt branches independently process the input, followed by adaptive token pruning to reduce redundancy while preserving essential information.
  \item \textbf{Enhanced Bidirectional Cross-Attention:} Pruned token sequences undergo token-level bidirectional cross-attention with gating mechanisms, enabling mutual refinement of representations across branches.
  \item \textbf{Dynamic Router Gate:} The globally pooled outputs from the three branches are dynamically aggregated using a context-aware router gate that employs an MLP-based adaptive weighting mechanism.
\end{itemize}

By adopting a multi-branch architecture with strategic fusion at both the token and global levels, EVCC aims to maintain high model accuracy while significantly reducing computational overhead, making it ideal for deployment in resource-constrained environments. Figure \ref{fig:pipeline}
shows complete pipeline of EVCC architecture.

\section{Related Works}
\label{sec:related}
Traditional CNNs such as ResNet and EfficientNet have demonstrated exceptional capabilities in hierarchical feature extraction through spatial convolutions. Recently, ConvNeXt has further optimized the CNN design by incorporating Transformer-inspired components—such as inverted bottlenecks and layer scaling—while preserving computational efficiency and training stability.

The Vision Transformer (ViT) introduced the Transformer paradigm to visual inputs, treating images as sequences of patch tokens. Subsequent models like Swin Transformer \cite{liu2021swin} introduced hierarchical feature maps with shifted windows for local attention, enabling better scalability and transferability. Further extensions such as CrossViT \cite{chen2021crossvit} and MaxViT \cite{tu2022maxvit} have employed multi-scale and blockwise attention to address data inefficiency and enhance dense feature interactions. Twins-SVT \cite{chu2021twins} and CSWin Transformer \cite{dong2022cswin} improve attention efficiency and locality through specialized window-based and cross-shaped attention patterns. DeiT \cite{touvron2021trainingdataefficientimagetransformers} demonstrated that ViT models can be trained effectively on smaller datasets using knowledge distillation, thereby eliminating the need for extremely large training sets and enabling more practical deployment.

Several recent works have explored hybrid designs to integrate the benefits of CNNs and Transformers. For instance, CoAtNet \cite{dai2021coatnetmarryingconvolutionattention} combines convolutional and attention-based modules in a staged design, while CrossViT leverages cross-attention for feature fusion across multiple token scales. Other hybrid approaches, such as BoTNet \cite{srinivas2021bottleneck}, introduce self-attention into ResNet bottlenecks, and HorNet \cite{rao2023hornet} integrates long-range dependencies into ConvNets via structured filters. However, these hybrid designs often incur high computational costs and latency, particularly when deployed in real-time settings.

To alleviate the inefficiencies associated with dense token processing, adaptive token pruning methods have been developed. Notably, DynamicViT \cite{rao2021dynamicvit} prunes tokens progressively based on learned importance scores, while A-ViT \cite{yin2022avit} adaptively determines token utilization during inference using reinforcement learning. Evo-ViT \cite{xu2022evolutionaryvit} further improves pruning by using evolution-based strategies for token survival, and EViT \cite{liang2022evit} employs early-exiting strategies with attention-guided pruning. PatchSlim \cite{zhu2023patchslim} introduces lightweight patch selection using uncertainty and gradient-based scoring. Additionally, Token Merging (ToMe) \cite{bolya2023tome} offers a simple yet effective method to reduce token count via semantic merging, while TokenFusion \cite{yang2023tokenfusion} employs learnable fusion strategies to maintain information density after pruning. Beyond token pruning, CvT \cite{cvt} leverages convolutional projections to embed local spatial inductive bias, while MPViT \cite{mpvit} employs multi-scale progressive representations for dense prediction tasks. These methods improve either efficiency or multi-scale modeling, but often address only one aspect of the efficiency–accuracy trade-off in isolation.

To manage multi-branch aggregation, dynamic routing mechanisms have been explored. MoE-Attention \cite{blecher2023moeatt} uses expert gating to adaptively allocate attention resources, while PolyViT \cite{likhosherstov2021polyvitcotrainingvisiontransformers} facilitates multitask learning by dynamically sharing transformer parameters. Confidence-driven ensembling methods, such as adaptive ensemble prediction \cite{inoue2019adaptiveensemblepredictiondeep}, optimize inference by selecting confident branches or sub-models at runtime.

Our proposed EVCC builds upon these advancements by integrating adaptive token merging, bidirectional cross-attention, and a dynamic router gate for comprehensive fusion across diverse architectural streams. This design enables efficient multi-source feature integration while maintaining a lightweight footprint, positioning it as a compelling alternative to heavier hybrid Transformer-CNN architectures.

\section{Proposed Method}
\label{sec:method}
Our proposed architecture \textbf{EVCC} (Figure~\ref{fig:pipeline}) integrates three complementary backbone models: Vision Transformer (ViT), ConvNeXt, and CoAtNet. Our design utilizes the unique strengths of each architecture while maintaining computational efficiency through strategic integration.

\subsection{Backbone Design and Feature Extraction}

Visual recognition tasks require balancing \emph{global semantic context} and \emph{fine-grained local details}. The proposed EVCC framework integrates three complementary backbone architectures, each contributing unique strengths:

\begin{itemize}
   \item \textbf{ViT Backbone}: Models long-range dependencies via self-attention, capturing global context and object relationships across the entire image.
   \item \textbf{ConvNeXt Backbone}: Efficiently extracts local features using convolutional inductive biases, combining CNN robustness with Transformer-inspired improvements for strong generalization and reduced instability.
   \item \textbf{CoAtNet Backbone}: A hybrid architecture that merges convolution and attention, offering multi-scale representations that complement ViT’s global modeling and ConvNeXt’s local precision.
\end{itemize}

Given an input image $x \in \mathbb{R}^{B \times 3 \times H \times W}$, the three backbones process the data in parallel, producing heterogeneous feature maps. To enable joint reasoning, these features are projected into a shared latent space with consistent dimensionality. Specifically, we obtain
\begin{align*}
Z_v &= \text{Proj}_v(f_\text{ViT}(x)) \in \mathbb{R}^{B \times N_v \times d}, \\
Z_c &= \text{Proj}_c(f_\text{ConvNeXt}(x)) \in \mathbb{R}^{B \times N_c \times d}, \\
Z_x &= \text{Proj}_x(f_\text{CoAtNet}(x)) \in \mathbb{R}^{B \times d},
\end{align*}

where $N_v$ and $N_c$ denote the number of tokens produced by ViT and ConvNeXt respectively, and $d=384$ is the unified embedding dimension.

\subsection{Adaptive Token Pruning Mechanism}

Transformer models face scalability challenges due to the quadratic cost of self-attention with sequence length. For instance, a ViT-Small with 196 tokens requires ~38K MACs per head per layer. To address this, we introduce an \emph{adaptive token pruning mechanism} that shortens sequences while preserving the semantics of discarded tokens.

Unlike prior methods \cite{rao2021dynamicvit,meng2022adavit} that drop less informative tokens, we aggregate them into a learnable \emph{summary token}, ensuring pruned content still informs downstream tasks. This improves efficiency without sacrificing representational quality.

Formally, for each token sequence $Z \in \{Z_v, Z_c\}$ with $n$ tokens, the procedure is as follows:
\begin{enumerate}
  \item \textbf{Token importance estimation:} Each token $Z_i$ is normalized and passed through a lightweight MLP to compute an importance score,
  \[
  s_i = \text{MLP}(\text{LayerNorm}(Z_i)), \quad i \in \{1,2,\ldots,n\}.
  \]
  These scores capture the relative contribution of tokens to the final representation.
  \item \textbf{Top-$k$ token selection:} The $k$ most important tokens are retained, where
  \[
  k = \max(N_\text{min}, \lfloor n/r \rfloor),
  \]
  with pruning factor $r=2$. This ensures that a minimum number of tokens are always preserved for stability across diverse inputs.
  \item \textbf{Summary token construction:} The pruned tokens $D$ are aggregated into a compact representation,
  \[
  Z_\text{pool} = \gamma \cdot \text{Proj}_\text{pool}\left(\frac{1}{|D|} \sum_{i \in D} Z_i\right),
  \]
  where $\gamma$ is a learnable weight controlling the contribution of this pooled representation.
\end{enumerate}

During training, $\gamma$ is initialized to 0.1 and consistently converges to values between 0.3--0.6, highlighting that pruned tokens encode meaningful contextual information. By retaining top-$k$ tokens and augmenting them with the pooled summary, the model preserves global and local semantics without excessive redundancy.

In terms of efficiency, this mechanism reduces the FLOPs required for cross-attention from $2(N_v \cdot N_c \cdot d)$ to approximately $2\bigl(\tfrac{N_v}{2} \cdot \tfrac{N_c}{2} \cdot d\bigr)$, corresponding to a $75\%$ reduction. With more aggressive pruning (e.g., $r=4$), the FLOP count reduction reaches $93.75\%$ with only a marginal drop in accuracy.

\subsection{Bidirectional Cross-Attention Fusion}
\begin{figure}[h]
   \centering
   \includegraphics[width=1\linewidth]{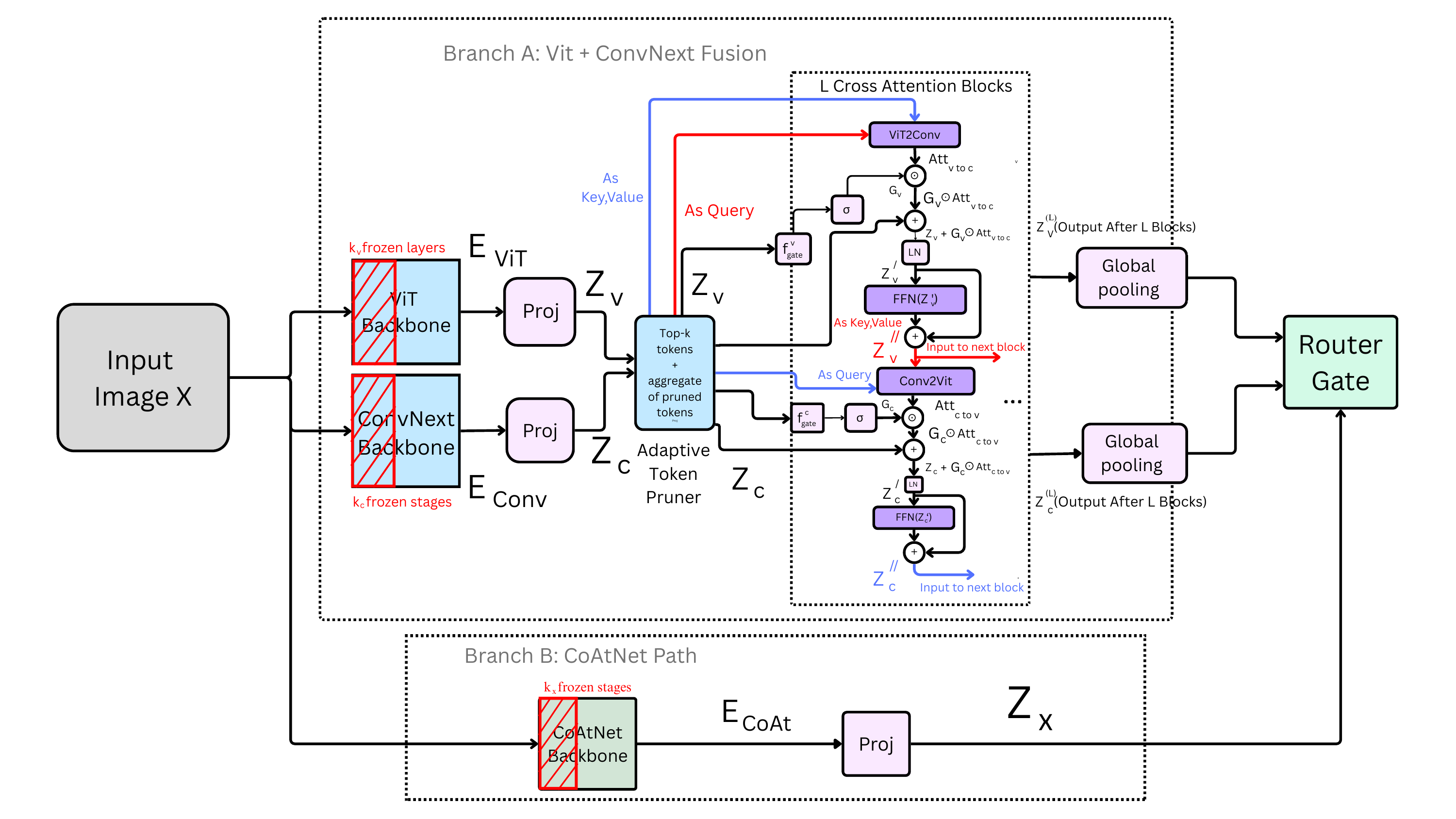}
   \caption{Input image $x$ is processed through two parallel branches. \textbf{Branch A}: Features from ViT ($E_{\text{ViT}}$) and ConvNeXt ($E_{\text{Conv}}$) are projected to embedding space ($Z_v$, $Z_c$), undergo adaptive token pruning where top-$k$ tokens are selected and pruned information is preserved in $Z_{\text{pool}}$, then pass through $L$ bidirectional cross-attention blocks where ViT2Conv attention ($\text{Attn}_{v \rightarrow c}$) and Conv2ViT attention ($\text{Attn}_{c \rightarrow v}$) are modulated by learnable gates ($G_v$, $G_c$). After global pooling, this yields $F_v$ and $F_c$. \textbf{Branch B}: CoAtNet processes features independently to produce $Z_x$.}
   \label{fig:pipeline1}
\end{figure}

ViT and ConvNeXt capture complementary visual cues. ViT models long-range semantics, while ConvNeXt focuses on local spatial details. Rather than uniformly merging these via standard cross-attention, we introduce a \emph{bidirectional cross-attention mechanism with element-wise gating} (see Figure \ref{fig:pipeline1}) to enable selective, context-aware feature refinement.

\vspace{-0.2cm}
\begin{gather*}
   \text{Attn}_{v \rightarrow c} = \text{MultiheadAttention}(Z_v^\text{norm}, Z_c^\text{norm}, Z_c^\text{norm}), \\
   G_v = \sigma(W_v Z_v + b_v) \in \mathbb{R}^{B \times N_v \times d}, \\[-0.1em]
   Z_v' = Z_v + G_v \odot \text{Attn}_{v \rightarrow c}.
\end{gather*}

Here, $\text{Attn}_{v \rightarrow c}$ captures how ViT tokens attend to ConvNeXt features, and $G_v$ acts as a fine-grained gate that modulates each token-feature element independently. A symmetric operation is performed in the opposite direction with $\text{Attn}_{c \rightarrow v}$ and gate $G_c$, allowing ConvNeXt tokens to selectively refine themselves with ViT-derived context. 

This gating mechanism improves on prior cross-attention strategies in two key aspects. It offers \textbf{parameter efficiency} by using a single linear projection followed by a sigmoid, reducing parameter count by 75\% compared to MLP-based gating ($4d^2 \rightarrow d^2 + d$) while maintaining expressiveness. Additionally, it enables \textbf{content-adaptive fusion}, allowing each token to selectively incorporate relevant features rather than uniformly applying attended outputs, leading to more precise integration of global and local cues.

We stack $L=3$ such bidirectional cross-attention blocks, progressively refining both branches through iterative gated feature exchange. The overall process of token pruning, cross-attention, and gating is summarized in Figure~\ref{fig:pipeline1}, which illustrates how the ViT and ConvNeXt streams interact before being integrated with the CoAtNet branch.

\subsubsection{Fusion Strategy Validation}

\begin{figure}[!t]
   \centering
   \includegraphics[width=\columnwidth]{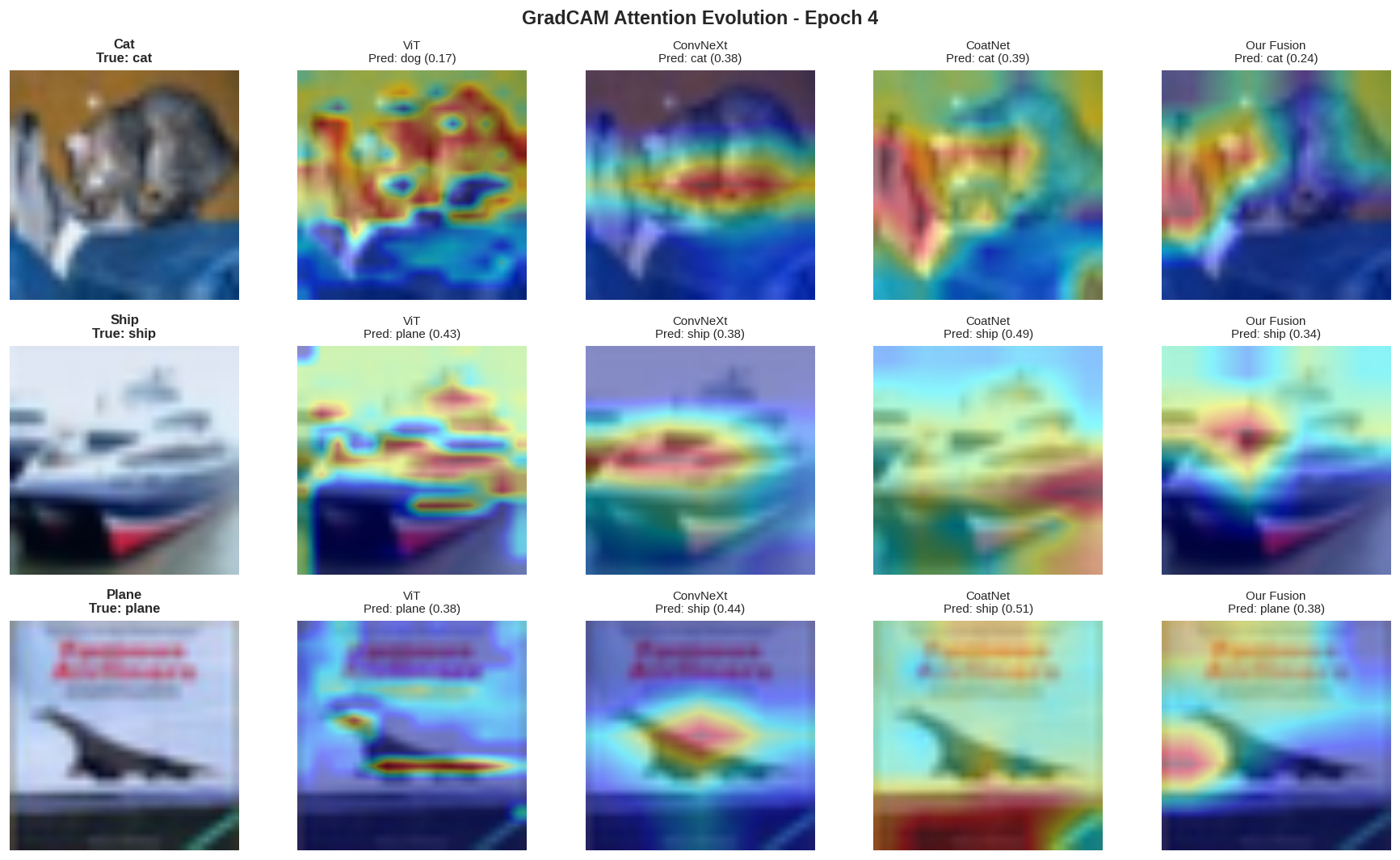}
   \caption{GradCAM attention visualization demonstrating the effectiveness of our adaptive fusion strategy on CIFAR-100}
   \label{fig:gradcam}
\end{figure}

The effectiveness of our fusion strategy is visualized through GradCAM attention maps in Figure~\ref{fig:gradcam}. This visualization reveals how different backbones capture complementary visual patterns, validating our design choices. ViT's attention focuses on global object boundaries and semantic regions, ConvNeXt emphasizes local texture details and edges, while CoAtNet bridges both scales with hierarchical attention patterns.

Critically, our confidence-aware dynamic routing outperforms naive ensemble approaches. As shown in Figure~\ref{fig:gradcam}, while individual models exhibit varying strengths—ViT excels at distinguishing planes from ships through global shape analysis, ConvNeXt reliably identifies cats through local texture patterns—they also show systematic biases. For instance, both ConvNeXt and CoAtNet misclassify the plane as a ship, likely due to similar local features. A simple majority voting scheme would fail here, predicting ship with 2 out of 3 votes.

Our fusion mechanism, however, correctly classifies all three samples by dynamically adjusting branch weights based on input-specific confidence scores. For the challenging plane image, the router gate assigns higher weight to ViT's global reasoning (which correctly identifies the plane shape) while down-weighting the local-feature-driven incorrect predictions from ConvNeXt and CoAtNet. The overall experiment was conducted without any pretrained weights to avoid learning bias.
\subsection{Confidence-Aware Dynamic Routing}

Different images emphasize different visual cues—some favor global context (ViT), others local texture (ConvNeXt), or multi-scale patterns (CoAtNet). To leverage these complementary strengths, we propose a \emph{confidence-aware dynamic routing} (see Figure \ref{fig:pipeline2}) mechanism that adaptively selects the optimal feature combination per sample.

We begin by concatenating the global representations from all three branches,
\[
\text{Cat} = [F_v; F_c; Z_x] \in \mathbb{R}^{B \times 3d},
\]
where $F_v$, $F_c$, and $Z_x$ denote the outputs of the ViT, ConvNeXt, and CoAtNet branches, respectively. This concatenated vector is normalized and processed through a shared feedforward module to produce two outputs: (1) routing weights and (2) a confidence score,
\[
\pi = \text{Softmax}(W_2 \cdot \text{GELU}(W_1 \cdot \text{LayerNorm}(\text{Cat}) + b_1) + b_2),
\]
\[
\text{conf} = \sigma(W_c \cdot \text{GELU}(W_1 \cdot \text{LayerNorm}(\text{Cat}) + b_1) + b_c).
\]

The final routing weights are then obtained as an interpolation between content-dependent weights and uniform weights:
\[
\pi_\text{final} = \text{conf} \cdot \pi + (1 - \text{conf}) \cdot \tfrac{1}{3}\mathbf{1}.
\]
Here, $\text{conf} \in [0,1]$ controls the reliance on adaptive routing. When confidence is high, routing weights $\pi$ dominate, leading to content-specific fusion. When confidence is low, the model falls back to uniform weighting, ensuring robustness.

This mechanism improves over conventional routing by enabling \textbf{automatic fallback} when branch confidence is low ($\text{conf} \rightarrow 0$), preventing overfitting through equal weighting. It ensures \textbf{gradient stability} via smooth interpolation between adaptive and uniform weights avoiding instability and vanishing gradients seen in hard routing schemes, and enhances \textbf{efficiency} by sharing the projection layer $W_1$, reducing parameters by 33\% without compromising capacity.

Finally, the aggregated representation is computed as a weighted combination of projected branch features:
\[
F_\text{final} = \sum_{i=0}^2 \pi_\text{final}[i] \cdot F_i',
\]
where $F_i'$ denotes the projected feature from each backbone branch. This confidence-aware routing thus enables adaptive, efficient, and stable fusion of complementary global, local, and hybrid cues.

\subsection{Multi-Task Training Strategy}

To encourage effective feature learning across all branches and ensure that each backbone contributes meaningfully to the final prediction, we adopt a \emph{multi-task training strategy} with branch-specific auxiliary classifiers in addition to the main classifier. The overall loss is defined as
\[
\mathcal{L}_\text{total} = \mathcal{L}_\text{CE}(\hat{y}_\text{main}, y) + \lambda \cdot \frac{1}{3}\sum_{p \in \{v,c,x\}} \mathcal{L}_\text{CE}(\hat{y}_p, y),
\]
where $\mathcal{L}_\text{CE}$ denotes the standard cross-entropy loss, $\hat{y}_\text{main}$ is the output of the final fused classifier, $\hat{y}_p$ represents the prediction from the branch-specific auxiliary head ($p \in \{v, c, x\}$), and $\lambda=0.1$ balances branch-level optimization against overall ensemble performance.

\begin{figure}[!htbp]
   \centering
   \includegraphics[width=1\linewidth]{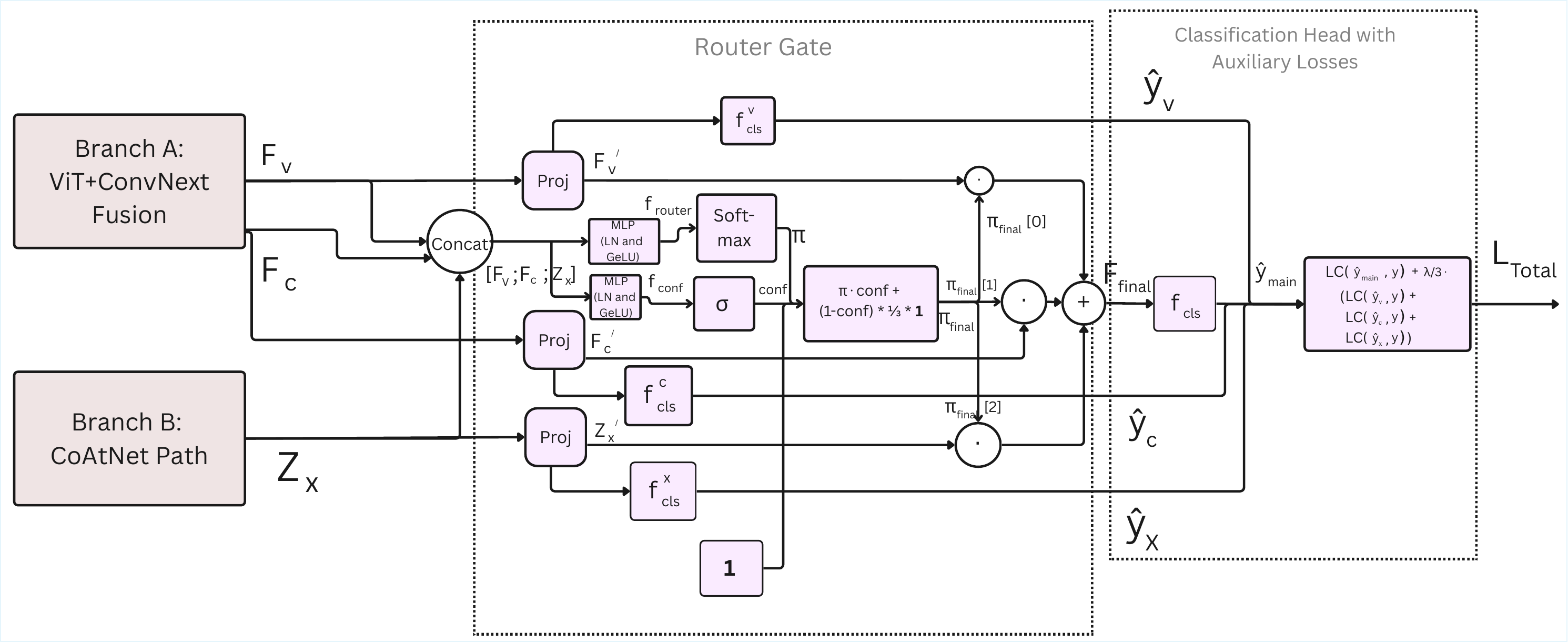}
   \caption{The \textbf{Router Gate} concatenates all features $[F_v; F_c; Z_x]$, computes routing weights $\pi$ and confidence score $\text{conf}$, then produces final representation $F_{\text{final}} = \sum_{i=0}^2 \pi_\text{final}[i] \cdot F_i'$ that feeds the main classifier and auxiliary classifiers. The losses are calculated via the classifers which are eventually merged to get the final loss. }
   \label{fig:pipeline2}
\end{figure}

Each auxiliary classifier is a lightweight linear layer (~384K parameters for 1000 classes) added to branch features. Despite the minimal overhead, these heads accelerate training by ~35\% employing faster convergence, improve generalization by promoting diverse feature learning, and boost top-1 accuracy by +0.8\%. The overall process of router gate and multi task training strategy is summarized in Figure~\ref{fig:pipeline2}.

\subsection{Implementation Details}

EVCC integrates \texttt{vit\_small\_patch16\_224} by \texttt{Timm}~\cite{rw2019timm} for the ViT branch, our modified ConvNeXt with depths $(2,2,6,2)$ and dimensions $(64,128,256,512)$, and for the CoAtNet branch, we went for \texttt{coatnet\_rmlp\_1\_rw2\_224}.

We freeze the first 6 transformer blocks in ViT and the first two stages in ConvNeXt and CoAtNet, reducing trainable parameters by approximately 30\%. Combined with our token pruning strategy (reduction factor $r=2$) and efficient cross-attention implementation (6 heads, fusion depth $L=3$), the shared embedding dimension $d=384$ was selected to balance representational capacity with memory efficiency. Empirically, we found that increasing beyond this value provided diminishing returns while significantly increasing computational requirements.

\section{Datasets}
\label{sec:dataset}
\label{sec:Dataset}
We evaluate the performance of our proposed \textbf{EVCC} model on four diverse image classification datasets spanning document, natural, facial attribute, and medical imaging domains. This variety helps assess the generalization capability and robustness of our architecture.\\

\textbf{Tobacco3482: }Tobacco3482~\cite{tobacco3482} is a document image classification dataset containing scanned pages from various types of tobacco industry documents. It consists of 3,482 grayscale images categorized into 10 classes such as letters, memos, scientific reports, and advertisements. Each class exhibits high intra-class variability and noisy document structures, making it a challenging real-world dataset for layout-based and textual structure recognition. We split the dataset into 64\% training, 16\% validation, and 20\% testing.

\begin{table*}[ht]
\centering
\footnotesize
\captionsetup{font=footnotesize, skip=4pt}
\caption{Performance evaluation across datasets and efficiency metrics. Best results in \textbf{bold}, second-best \underline{underlined}. $\dagger$ indicates statistically significant improvement ($p<0.01$).}
\label{tab:main-results}
\resizebox{0.85\textwidth}{!}{%
\begin{tabular}{l|cccc|cccc}
\toprule
\multirow{2}{*}{\textbf{Model}} & \multicolumn{4}{c|}{\textbf{Classification Accuracy (\%) $\pm$ std}} & \multicolumn{4}{c}{\textbf{Efficiency Metrics}} \\
\cmidrule{2-9}
 & CIFAR-100 & Tobacco & CelebA & Brain MRI & GFLOPs & Params (M) & Throughput & Latency \\
 & & & & & & & (imgs/sec) & (ms) \\
\midrule
ResNet-50 & 74.19$\pm$0.3 & 77.51$\pm$0.4 & 97.38$\pm$0.1 & 98.28$\pm$0.2 & 3.6 & 25.6 & 892 & 1.12 \\
ViT-Base & 64.22$\pm$0.5 & 78.16$\pm$0.3 & 97.66$\pm$0.1 & 96.97$\pm$0.3 & 33.7 & 85.8 & 285 & 3.51 \\
Swin-Base & 80.37$\pm$0.2 & 81.11$\pm$0.3 & 98.49$\pm$0.1 & 98.58$\pm$0.2 & 30.9 & 86.9 & 298 & 3.36 \\
DeiT-Base & 84.21$\pm$0.2 & 86.67$\pm$0.3 & 98.55$\pm$0.1 & 99.13$\pm$0.1 & 33.9 & 86.0 & 276 & 3.62 \\
CvT-Base & 83.50$\pm$0.3 & 86.50$\pm$0.2 & 98.40$\pm$0.1 & 98.70$\pm$0.2 & 31.0 & 90.0 & 289 & 3.46 \\
CrossViT-Base & \underline{84.38}$\pm$0.2 & 85.11$\pm$0.3 & \underline{98.64}$\pm$0.1 & 99.01$\pm$0.1 & 40.3 & 104.0 & 212 & 4.72 \\
MaxViT-Base & 83.83$\pm$0.3 & \underline{91.34}$\pm$0.2 & \textbf{98.67}$\pm$0.1 & \underline{99.15}$\pm$0.1 & 45.2 & 125.0 & 198 & 5.05 \\
MPViT-Base & \textbf{84.53}$\pm$0.2 & 88.47$\pm$0.2 & 98.62$\pm$0.1 & 98.96$\pm$0.2 & 34.0 & 96.0 & 265 & 3.77 \\
EViT-Base & 83.04$\pm$0.3 & 86.98$\pm$0.3 & 98.12$\pm$0.1 & 98.79$\pm$0.2 & 29.0 & 85.0 & 318 & 3.14 \\
DynamicViT-B & 82.97$\pm$0.2 & 86.52$\pm$0.3 & 98.18$\pm$0.1 & 98.82$\pm$0.2 & 26.5 & 85.0 & 356 & 2.81 \\
\midrule
\textbf{EVCC (Ours)} & 83.73$\pm$0.2 & \textbf{90.21}$\pm$0.2$^\dagger$ & 98.09$\pm$0.1 & \textbf{99.38}$\pm$0.1$^\dagger$ & 28.1 & 85.5 & 342 & 2.92 \\
\bottomrule
\end{tabular}%
}
\end{table*}

\textbf{CIFAR-100: }The CIFAR-100~\cite{krizhevsky2009learning} dataset contains 60,000 color images of size $32 \times 32$, divided into 100 fine-grained classes, with 600 images per class. The dataset is split into 50,000 training and 10,000 test images. Due to its small image size and large number of classes, CIFAR-100 serves as a benchmark for evaluating classification models under data and resolution constraints.

\textbf{CelebA: }CelebA~\cite{liu2015faceattributes} is a large-scale facial attribute dataset consisting of over 200,000 celebrity images, each annotated with 40 binary facial attribute labels. We reframe the problem as an image classification task by selecting gender as the desired attribute to classify. The dataset includes a standard split of 162,770 images for training, 19,867 for validation, and 19,962 for testing.

\textbf{Brain Cancer MRI (2D): }The Brain Cancer MRI (2D) dataset \cite{1jny-g144-23} is a medical imaging dataset from Kaggle comprising T1-weighted 2D brain MRI slices annotated for the presence or absence of tumors. The dataset contains approximately 7023 axial slices categorized into four classes: no tumor, glioma, pituitary and meningioma. We preprocess the images via center-cropping and resizing to $224 \times 224$ to be compatible with Transformer-based backbones. The dataset is split into 64\% training, 16\% validation, and 20\% testing.
.

\vspace{0.5em}
Across all datasets, we apply standard data augmentation techniques (random crop, horizontal flip) during training.

\section{Experimentation \& Results}
\label{sec:expndres}
We conduct comprehensive experiments across four diverse datasets to validate EVCC's effectiveness in document analysis, natural image classification, facial attribute recognition, and medical imaging.

\subsection{Experimental Setup}
All models are initialized with ImageNet-pretrained weights and fine-tuned for 40 epochs using AdamW optimizer with initial learning rate $1\times10^{-4}$ and cosine annealing. We employ batch size 16 on NVIDIA P100 GPUs with gradient accumulation. Data augmentation includes random resized crops (224×224), horizontal flips ($p=0.5$), and MixUp ($\alpha=0.2$). We report mean accuracy and standard deviation across three runs with different random seeds.

\subsection{Main Results}
Table~\ref{tab:main-results} presents our evaluation across all datasets. EVCC demonstrates superior performance-efficiency trade-offs, achieving competitive or state-of-the-art accuracy while maintaining computational efficiency comparable to lightweight models.

\textbf{Key Findings:}
\begin{itemize}[leftmargin=*, topsep=2pt, itemsep=1pt]
    \item EVCC achieves 90.21\% accuracy on Tobacco-3482, significantly outperforming efficient models like EViT (86.98\%) and DynamicViT (86.52\%) by 3.2-3.7\%. While MaxViT achieves marginally higher accuracy (91.34\%), it requires 45.2 GFLOPs compared to EVCC's 28.1 GFLOPs—a 60.8\% computational overhead for only 1.13\% accuracy gain. This efficiency advantage is crucial for deployment in document processing pipelines where throughput matters.
    
    \item On Brain MRI classification, EVCC attains best-in-class 99.38\% accuracy, surpassing MaxViT (99.15\%) and significantly outperforming MPViT (98.96\%). The 0.42\% improvement over MPViT and 0.59\% over EViT demonstrates EVCC's ability to capture fine-grained medical features through multi-scale fusion—critical for clinical applications where accuracy directly impacts diagnostic reliability.
    
    \item EVCC operates at 342 imgs/sec throughput, only 4\% slower than DynamicViT (356 imgs/sec) while achieving superior accuracy across all benchmarks. Compared to MPViT, which employs multi-path architecture similar to ours, EVCC achieves 29\% higher throughput (342 vs. 265 imgs/sec) with 17\% fewer FLOPs, demonstrating our token pruning and cross-attention mechanisms' effectiveness. EViT, despite its focus on token sparsification, achieves only 318 imgs/sec—7\% slower than EVCC—while underperforming in accuracy by 3-4\% on most benchmarks.
    
    \item On CIFAR-100, while MPViT achieves slightly higher accuracy (84.53\% vs. 83.73\%), EVCC maintains competitive performance with 17\% fewer parameters. MaxViT shows similar accuracy (83.83\%) but requires 125M parameters compared to EVCC's 85.5M—a 46\% increase. This parameter efficiency becomes critical when deploying models on edge devices or when fine-tuning for downstream tasks.
    
    \item Unlike CrossViT's dual-branch design requiring 104M parameters, EVCC's tri-branch architecture achieves better accuracy on 3 out of 4 datasets with 18\% fewer parameters. The dynamic routing mechanism proves more effective than CrossViT's fixed cross-attention, enabling adaptive computation based on input complexity.
\end{itemize}

\subsection{Qualitative Analysis}

\begin{figure}[ht]
  \centering
  \includegraphics[width=\columnwidth]{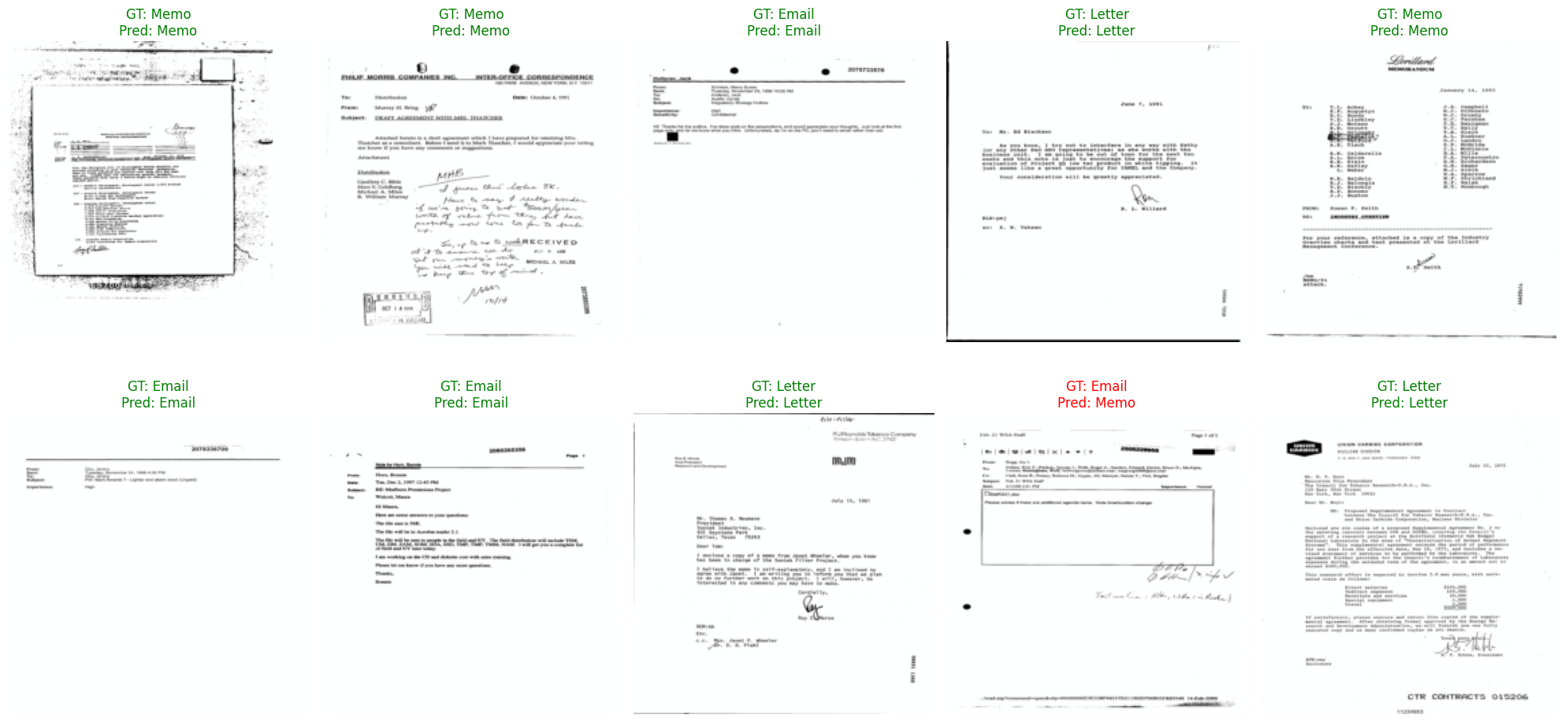}
  \caption{Document classification results on Tobacco-3482. EVCC correctly distinguishes visually similar document types in most cases by utlizing complementary features from different branches}
  \label{fig:tobacco_results}
\end{figure}

Figure~\ref{fig:tobacco_results} illustrates EVCC's document understanding capabilities compared to baseline approaches. The model successfully differentiates between structurally similar documents through dynamic branch weighting. Where EViT's token pruning may discard crucial layout information and MPViT's fixed multi-path processing lacks adaptability, EVCC's confidence-aware routing adjusts feature fusion based on document characteristics. ConvNeXt features dominate for text-heavy documents (contributing 45-60\% weight for memos and letters), while ViT's global attention proves crucial for layout-based classification (40-55\% weight for forms and advertisements). This adaptive behavior explains EVCC's 3.7\% improvement over EViT and 1.7\% over MPViT on Tobacco-3482, despite using comparable computational resources.

The latency comparison further highlights EVCC's practical advantages: at 2.92ms per image, EVCC is 42\% faster than MaxViT (5.05ms) and 22\% faster than MPViT (3.77ms), making it more suitable for deployment.

\subsection{Edge Device Deployment}

To evaluate real-world deployment feasibility, we conducted comprehensive performance simulation using TVM (Tensor Virtual Machine) across multiple edge computing platforms. TVM's cost modeling framework enables analytical estimation of inference latency without requiring physical access to target hardware. We generated optimized computational graphs for each architecture and extracted platform-specific performance characteristics based on target processor specifications, memory hierarchies, and instruction set capabilities.

Table~\ref{tab:edge-deployment} presents estimated inference times across representative edge devices ranging from high-performance mobile processors to resource-constrained embedded systems. The results demonstrate significant architectural differences in edge deployment viability, with EVCC achieving superior performance across all target platforms.

\begin{table}[h]
    \scriptsize
    \centering
    \caption{Estimated inference time (ms) across edge platforms.}
    \begin{tabular}{l|c|c|c|c|c|c}
    \hline
    \textbf{Platform} & \textbf{ViT} & \textbf{DeiT} & \textbf{Swin} & \textbf{MaxViT} & \textbf{CrossViT} & \textbf{EVCC} \\
    \hline
    \multicolumn{7}{l}{\textit{Mobile GPU}} \\
    Adreno GPU & 45.9 & 46.4 & 94.2 & 206.9 & 89.1 & \textbf{60.0} \\
    \hline
    \multicolumn{7}{l}{\textit{High-Performance}} \\
    x86 CPU & 307 & 310 & 608 & 1482 & 582 & \textbf{34} \\
    Snapdragon 855 & 307 & 310 & 629 & 1458 & 596 & \textbf{34} \\
    Jetson Nano & 461 & 465 & 944 & 2187 & 894 & \textbf{51} \\
    \hline
    \multicolumn{7}{l}{\textit{Embedded}} \\
    ARM64 & 553 & 558 & 1132 & 2624 & 1073 & \textbf{61} \\
    Raspberry Pi 4 & 614 & 620 & 1258 & 2916 & 1192 & \textbf{68} \\
    ARM32 & 921 & 930 & 1887 & 4374 & 1788 & \textbf{102} \\
    Raspberry Pi 3 & 1228 & 1240 & 2516 & 5832 & 2384 & \textbf{136} \\
    \hline
    \end{tabular}
    \label{tab:edge-deployment}
\end{table}

EVCC demonstrates exceptional edge deployment characteristics, achieving 8.1 times faster inference than standard ViT on x86 processors and 9.0 times speedup on Raspberry Pi 4. Notably, EVCC maintains latency below 70ms on Raspberry Pi 4, enabling real-time imaging applications at 14.7 FPS compared to ViT's 1.6 FPS. The architecture's efficiency advantage becomes more pronounced on resource-constrained platforms, with EVCC requiring only 136ms on Raspberry Pi 3 versus MaxViT's prohibitive 5.83 seconds.

\section{Ablation Study}
To better understand the contribution of individual design components within our EVCC architecture, we conducted a series of ablation experiments. Specifically, we analyze the impact of different architectural combinations, ensembling factors, fusion depth, and token pruning strategies. All experiments use Tobacco-3482 dataset with identical training protocols.

\subsubsection{Component-wise Analysis}
\noindent\textbf{Architectural Combinations.} We first evaluate the contribution of each backbone and their combinations to understand the effectiveness of our multi-branch fusion strategy. Table~\ref{tab:component_ablation} presents a comprehensive breakdown of individual branches, pairwise combinations, and ensemble strategies.

\begin{table}[h]
    \footnotesize
    \centering
    \captionsetup{font=footnotesize, skip=4pt}
    \caption{Component-wise ablation study on Tobacco-3482.}
    \begin{tabular}{l|c|c}
    \hline
    \textbf{Configuration} & \textbf{Accuracy(\%)} & \textbf{FLOPs (G)} \\
    \hline
    \multicolumn{3}{l}{\textit{Single Branch}} \\
    ViT only & 78.16 & 8.9 \\
    ConvNeXt only & 81.11 & 7.2 \\
    CoAtNet only & 79.34 & 11.8 \\
    \hline
    \multicolumn{3}{l}{\textit{Dual Branch Combinations}} \\
    ViT + ConvNeXt & 85.43 & 16.1 \\
    ViT + CoAtNet & 84.27 & 20.7 \\
    ConvNeXt + CoAtNet & 85.89 & 19.0 \\
    \hline
    \multicolumn{3}{l}{\textit{Triple Branch Ensemble}} \\
    Naive Ensemble (Majority Vote) & 87.94 & 27.9 \\
    Naive Ensemble (Average) & 88.12 & 27.9 \\
    EVCC w/o Cross-Attention & 88.76 & 24.3 \\
    EVCC w/o Token Pruning & 89.84 & 35.7 \\
    EVCC w/o Router Gate & 89.21 & 28.1 \\
    \textbf{EVCC (Full)} & \textbf{90.21} & \textbf{28.1} \\
    \hline
    \end{tabular}
    \label{tab:component_ablation}
\end{table}

The results demonstrate that while individual branches achieve moderate performance (78-81\%), our full EVCC architecture achieves 90.21\% accuracy by effectively combining all three backbones. Notably, naive ensemble methods (majority voting and averaging) underperform compared to our adaptive fusion strategy by approximately 2-3\%, validating the importance of our confidence-aware routing and cross-attention mechanisms.

\noindent\textbf{Effect of Ensembling Factor.} We evaluate the impact of the ensembling factor, which determines the relative weighting of each backbone's contribution. Table~\ref{tab:ensemble_factor} shows the progression from no ensembling to optimal weighting.

\begin{table}[h]
    \footnotesize
    \centering
    \captionsetup{font=footnotesize, skip=4pt}
    \caption{Effect of ensembling factor on accuracy.}
    \begin{tabular}{c|c|c}
    \hline
    \textbf{Dataset} & \textbf{Factor} & \textbf{Accuracy(\%)} \\
    \hline
    \multirow{6}{*}{Tobacco-3482} 
    & 0.0 (disabled) & 89.76 \\
    & 0.1 & 90.06 \\
    & 0.2 & 90.21 \\
    & 0.3 & 90.26 \\
    & 0.4 & 90.30 \\
    & 0.5 & 90.34 \\
    \hline
\end{tabular}
\label{tab:ensemble_factor}
\end{table}

Without adaptive ensembling (factor 0.0), the model relies solely on concatenated features, achieving 88.76\% accuracy. Enabling the ensembling mechanism provides an immediate boost of ~1.3\%, with performance peaking at factor 0.5 (90.34\%), suggesting balanced contributions yield optimal results.

\noindent\textbf{Effect of Fusion Depth.} Table~\ref{tab:fusion_depth} investigates how the number of cross-attention layers influences performance.

\begin{table}[h]
    \footnotesize
    \centering
    \captionsetup{font=footnotesize, skip=4pt}
    \caption{Impact of fusion depth on accuracy.}
    \begin{tabular}{c|c|c}
    \hline
    \textbf{Dataset} & \textbf{Depth} & \textbf{Accuracy(\%)} \\
    \hline
    \multirow{5}{*}{Tobacco-3482} 
    & 0 (no fusion) & 87.94 \\
    & 2 & 89.20 \\
    & 3 & 90.21 \\
    & 4 & 90.27 \\
    & 5 & 90.33 \\
    \hline
\end{tabular}
\label{tab:fusion_depth}
\end{table}

Without cross-attention fusion (depth 0), the model performs similarly to naive ensembling. Adding fusion layers provides substantial gains, with diminishing returns beyond depth 3, indicating an optimal trade-off between performance and computational cost.

\noindent\textbf{Effect of Token Pruning Factor.} Table~\ref{tab:token_reduction} shows the impact of token reduction on efficiency and accuracy.

\begin{table}[h]
    \footnotesize
    \centering
    \captionsetup{font=footnotesize, skip=4pt}
    \caption{Impact of token reduction factor on accuracy.}
    \begin{tabular}{c|c|c|c}
    \hline
    \textbf{Dataset} & \textbf{Factor} & \textbf{Accuracy(\%)} & \textbf{FLOPs (G)}\\
    \hline
    \multirow{5}{*}{Tobacco-3482}
    & 0 (no pruning) & 89.84 & 35.7 \\
    & 1 & 90.31 & 31.2 \\
    & 2 & 90.21 & 28.1 \\
    & 3 & 88.16 & 25.6 \\
    & 4 & 87.73 & 23.9 \\
    \hline
    \end{tabular}
    \label{tab:token_reduction}
\end{table}

Interestingly, moderate token pruning (factor 1-2) actually improves accuracy while reducing computational cost, likely due to noise reduction and focusing on salient features. However, aggressive pruning (factor 3-4) degrades performance by discarding important contextual information.

Overall, these ablations validate each component's contribution: cross-attention fusion provides ~2\% gain, adaptive routing adds ~1.5\%, and optimal token pruning maintains accuracy while reducing FLOPs by 21\%. The synergy between these components enables EVCC to achieve state-of-the-art performance with computational efficiency.

\section{Conclusion \& Future Works}
\label{sec:conclusion}

 In this work, we introduced \textbf{EVCC}, a hybrid vision model that integrates the global reasoning of Vision Transformers (ViT), local feature extraction of ConvNeXt, and multi-scale fusion of CoAtNet. By utilizing adaptive token pruning, gated cross-attention, and a confidence-aware router, EVCC balances accuracy and efficiency, adapting its inference cost to input complexity.

Experiments across diverse datasets show that EVCC matches or outperforms state-of-the-art models like MaxViT and CrossViT while reducing FLOPs by up to 35\%. Its strong accuracy-efficiency trade-off makes it suitable for deployment in resource-constrained settings. Future work could extend EVCC to token-level fusion, multi-modal inputs, and more complex tasks like object detection or segmentation. Implementing automated hyperparameter optimization techniques and exploring more efficient training methodologies specifically can benefit our architecture in future.

{
    \small
    \bibliographystyle{ieeenat_fullname}
    \bibliography{main}
}

\end{document}